\documentclass[conference]{IEEEtran}
\usepackage{times}

\usepackage{graphicx}
\usepackage{amsmath}
\usepackage{amssymb}
\usepackage{booktabs}
\usepackage{xcolor}
\usepackage{multirow}
\usepackage{bbm}
\usepackage{multicol,epigraph,wrapfig}
\usepackage{algorithm, algorithmic}
\usepackage{graphics,graphicx,caption,float,subcaption,booktabs,xcolor,multirow,array,color,ifthen,tabu,colortbl,dblfloatfix,url,xparse,mathtools,patchcmd,algorithm,algorithmic,amssymb,xspace,nicefrac,microtype,amsmath,amsfonts,bm,ragged2e,tikz,stackengine,etoolbox,xpatch,enumerate,xstring,setspace,tabularx,makecell,changepage,cuted,titlesec,wrapfig,tcolorbox, sidecap}

\usepackage[para]{footmisc}
\usepackage[pagebackref=true,breaklinks=true,colorlinks=true,bookmarks=false,citecolor=blue]{hyperref}

\usepackage[numbers]{natbib}
\usepackage{multicol}

\makeatletter
\newcommand{\removeParBefore}{\ifvmode\vspace*{-\baselineskip}\setlength{\parskip}{0ex}\fi}
\newcommand{\removeParAfter}{\@ifnextchar\par\@gobble\relax}
\newcommand{\eq}{\begingroup\removeParBefore\endlinechar=32 \eqinner}
\newcommand{\eqinner}[2][aligned]{\endlinechar=32%
\begin{gather}\begin{#1}#2\end{#1}\end{gather}\endgroup\removeParAfter}
\makeatother

\newcommand{\ours}{\mbox{{SWIM}}\xspace}
\begin{document}
\title{Structured World Models from Human Videos}  

\author{Russell Mendonca$^\star$ $\qquad$Shikhar Bahl$^\star$ $\qquad$Deepak Pathak\\ Carnegie Mellon University}

\maketitle

\begin{abstract}
We tackle the problem of learning complex, general behaviors directly in the real world. We propose an approach for robots to efficiently learn manipulation skills using only a handful of real-world interaction trajectories from \textit{many different settings}. Inspired by the success of learning from large-scale datasets in the fields of computer vision and natural language, our belief is that in order to efficiently learn, a robot must be able to leverage internet-scale, human video data. Humans interact with the world in many interesting ways, which can allow a robot to not only build an understanding of useful \textit{actions} and \textit{affordances} but also how these actions affect the world for manipulation. Our approach builds a structured, human-centric action space grounded in visual affordances learned from human videos. Further, we train a \textit{world model} on human videos and fine-tune on a small amount of robot interaction data without any task supervision. We show that this approach of affordance-space world models enables different robots to learn various manipulation skills in complex settings, in under 30 minutes of interaction. Videos can be found at \url{https://human-world-model.github.io}
\end{abstract}

\IEEEpeerreviewmaketitle
\section{Introduction}

A truly useful home robot should be general purpose, able to perform arbitrary manipulation tasks, and \textit{get better} at performing new ones as it obtains more experience. How can we build such generalist agents? The current paradigm in robot learning is to train a policy, in simulation or directly in the real world, with engineered rewards or demonstrations directly constructed for the environment. While this has shown successes in lab-based tasks  \cite{kalashnikov2018qt, pinto2015supersizing, levineFDA15}, learning is heavily dependent on the structure of the reward. This is not scalable as it is very challenging to \textit{transfer} to new tasks, with different objectives. Often, it is also difficult to obtain ground truth objectives for a task in the real world. For a robot to succeed in the wild, it must not only learn many tasks at a time but also \textit{get better} as it sees more data. How can we build an agent that can take advantage of large-scale experience and multi-task data? 

We aim to build \textit{world models} to tackle this challenge. One key observation is that there is commonality between many tasks performed by humans on a daily basis. Even across diverse settings, the environment dynamics and physics share a similar structure. Learning a single \emph{joint} world model, that predicts the future consequences of actions across diverse tasks can thus enable agents to extract this shared structure.

While world models enable learning from inter-task data, they require action information to make predictions about the future. Furthermore, for planning in an environment, the actions need to be relevant to the particular robot. Consequently, world models for robotics have mostly been trained only on data collected directly by a robot \cite{ebert2018foresight, wu2021example, lee2018stochastic, dasari2019robonet, ebert2021bridge}. However, the quantity of this data is limited, which is very expensive and cumbersome to collect in the real world. Thus, the benefits of using large datasets as seen in other machine learning areas such as computer vision and language \cite{gpt3, clip} have not been realized for robotics, as no such dataset exists for robotics. However, there is an abundance of \textit{human videos}, performing a very large set of tasks, on the internet. Is there a way to leverage this abundant data to learn world models for robotics, that will enable the robot to predict the consequences of its \emph{actions} in any environment, enabling general-purpose learning?

\begin{figure}[t!]
    \centering
    \includegraphics[width=\linewidth]{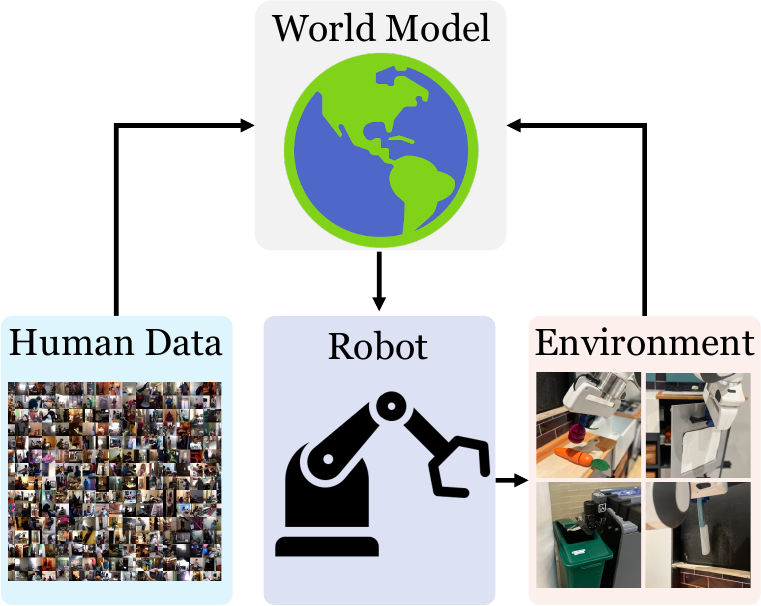}
    \caption{ \small We present \ours, an approach for learning manipulation tasks in the real world with only a handful of trajectories and only 30 min of real-world sampling.}
    \label{fig:teaser}
    \vspace{-0.2in}
\end{figure}

\begin{figure*}[t!]
    \centering
    \includegraphics[width=\linewidth]{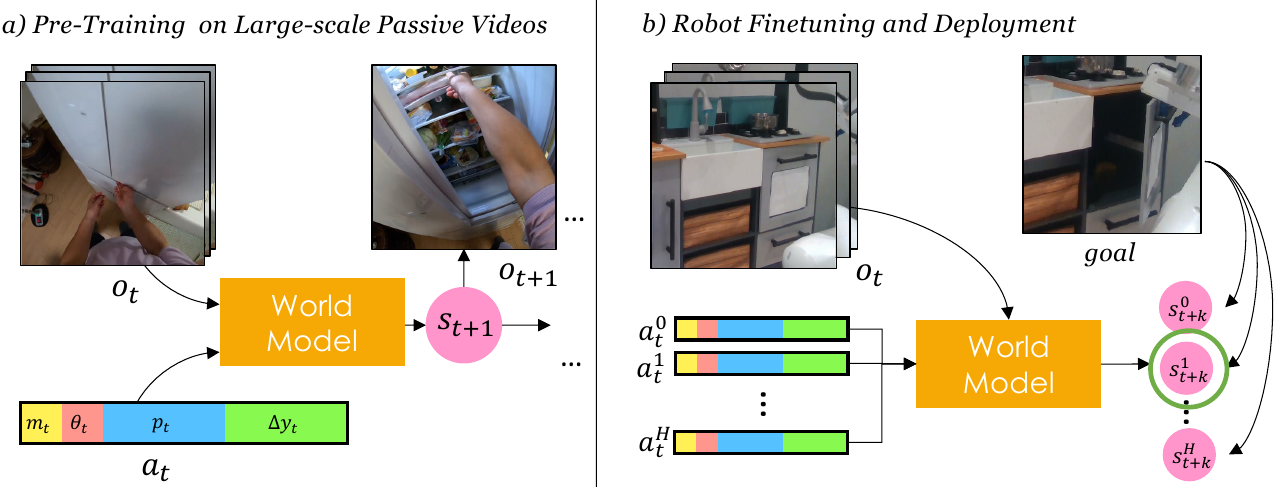}
    \caption{ \small Overview of \ours. We first pre-train the world model on a large set of human videos. We finetune this on many robot tasks, in an unsupervised manner, and deploy at test-time in the real world to achieve a given goal. Videos can be found at \texttt{\url{https://human-world-model.github.io}}}
    \label{fig:method}
\end{figure*}

 Due to the large morphology gap between robots and humans, it is challenging to obtain actions from human videos. Thus, previous approaches have mostly focused on learning visual representation features \cite{r3m, mvp} from observations alone.  Using internet human videos to train robots requires us to define an action space that is applicable both in the human video domain and for robots. Consider the task of picking up a mug. To perform this task, the low-level signals sent to a person's arm compared to that of a robot would be completely different, and so predictive models in low-level joint space will not transfer well. If the action space instead required predicting the target pose and orientation of the mug handle,  with low-level control abstracted away, then target poses used by humans could be utilized by robots as well. Thus, we learn \emph{high-level} structured action spaces that are morphology invariant. For manipulation tasks, predicting a grasp location and post-grasp waypoints is an effective action space since it encourages object interaction. We can train visual affordance networks that produce these locations given videos leveraging techniques in computer vision~\cite{hoi, hap, 100doh, hotspots, vrb}.

 In this paper we propose \textbf{S}tructured \textbf{W}orld \textbf{M}odels for \textbf{I}ntentionality (\textbf{\ours}),
 which utilizes large-scale internet data to train world models for robotics using structured action spaces. Training the world model in the common high-level structured action space allows it to capture how human hands interact with objects when trying to grasp and manipulate them. This model can then be fine-tuned for robotics settings with only a handful of real-world interaction trajectories. This is because the world model can leverage the actionable representations it was pre-trained with due to the similarity in how the human hands from video data and robot grippers interact with the world.
 Furthermore, these interaction trajectories for fine-tuning do not require any task supervision and can be obtained simply by executing the visual affordance actions. We note that both pre-training on human videos and finetuning the world model on robot data do not make any assumption on rewards, and this \emph{unsupervised} setting allows us to utilize data relevant for different tasks. This allows the robot to train a \emph{single} world model on all the data, thus enabling us to train generalist agents. In our experiments, we show that we can train such joint world models through two distinct robot systems operating in real-world environments. Finally, we can deploy the fine-tuned world model to perform tasks of interest by specifying a goal image. The world model then plans in the affordance action space to find a sequence of actions to manipulate objects as required by the task. 

 To summarize, \ours trains world models for robot control and consists of three stages: 1) Leveraging internet videos of human interactions for pre-training the model, 2) Finetuning the model to the robot setting using reward-free data, 3) Planning through the model to achieve goals. We evaluate this framework on two robot systems -- a Franka Arm, and a Hello Stretch robot. \ours is able to learn directly, is trained on data from multiple settings and gets better with data from more tasks.  We perform a large-scale study across multiple environments and robots and find that \ours achieves higher success ($\sim$ 2X) than prior approaches while being very sample efficient, requiring less than 30 minutes of real-world interaction data. 
 
\section{Related Work}

\noindent\textbf{Efficient Real World Robot Learning$\quad$} Deploying learning-driven approaches on hardware is challenging and requires either large engineering efforts to collect demonstrations \cite{brohan2022rt, jang2022bc}, many hours of autonomous interactions \cite{kalashnikov2018qt, kalashnikov2021mt}, or simulations \cite{andrychowicz2020learning, tobin2018domain, kumar2021rma}. A major constraint of continuous control is the extremely large action space. Prior methods have focused on reducing this search space by using skills or options in a hierarchical manner \cite{dalal2021accelerating, pastor2011skill, bacon2017option, sutton1999temporal, daniel2016hreps, parisi2015tetherball, daniel2016hreps}, physical inductive biases \cite{mulling2013learning, peters2003reinforcement, kormushev2010robot, stulp2012sequences, kober2009learning, bahl2020neural, vices2019martin}. It is also possible to visually ground the action space, by parameterizing each observed location by a 2D \cite{zeng2020transporter, zeng2018learning, shridhar2022cliport, james2022coarse} or 3D \cite{shridhar2022perceiver} action. While these can speed up learning, we find that our structured action space, based on human-centric visual affordances allows us to not only perform manipulation efficiently but also leverage out-of-domain human/internet videos. 

\noindent\textbf{Model-based learning$\quad$} To tackle the sample efficiency problem in robot learning, prior methods have proposed learning dynamics models, which can later be used to optimize the policy \cite{deisenroth2011pilco, deisenroth2013policy, hafner2018planet, chua2018pets, nagabandi2017mbmf, nagabandi-pddm}. Such approaches mostly operate and learn in state space, which tends to be low dimensional. In order to deal with the highly complex visual observations from real-world settings, prior methods have used \textit{World Models} \cite{ha2018worldmodels}, which capture dynamics of the \textit{agent} and its \textit{environment}. Such models can plan in image space \cite{finn2017foresight, ebert2018foresight} or fully in imagination space \cite{weber2017i2a, hafner2019dreamer, hafner2018learning, lee2019slac, sekar2020plan2explore, mendonca2021discovering}. Such world models have also shown to be useful for real robot tasks \cite{alan, daydreamer}. In this paper, we argue that world models can be helpful in modeling the real world, especially if they can understand how the environment will behave at a high level and model the intentions of the agent. 

\noindent\textbf{Visual and Action Pre-Training for Robotics$\quad$} In order to learn more generalizable and actionable representations, prior methods have learned visual encoders from large-scale human video data, either via video-language contrastive learning \cite{r3m} or through inpainting masked patches \cite{mvp, realmvp}. These representations have been shown to be useful for dynamics models as well \cite{hansen2022modem}. Such approaches focus on the visual complexity of the world but do not encode any behavior information. Some works have incorporated low-level actions from human videos into the learning loop \cite{dexvip, qin2022dexmv, arunachalam2022dexterous, shaw2022video}, but these are fixed for a specific morphology and use a direct mapping to the robot. In contrast, our approach is able to learn a world model from human videos, incorporating action information, and works in multiple settings. 

\begin{figure}[t!]
    \centering
    \includegraphics[width=\linewidth]{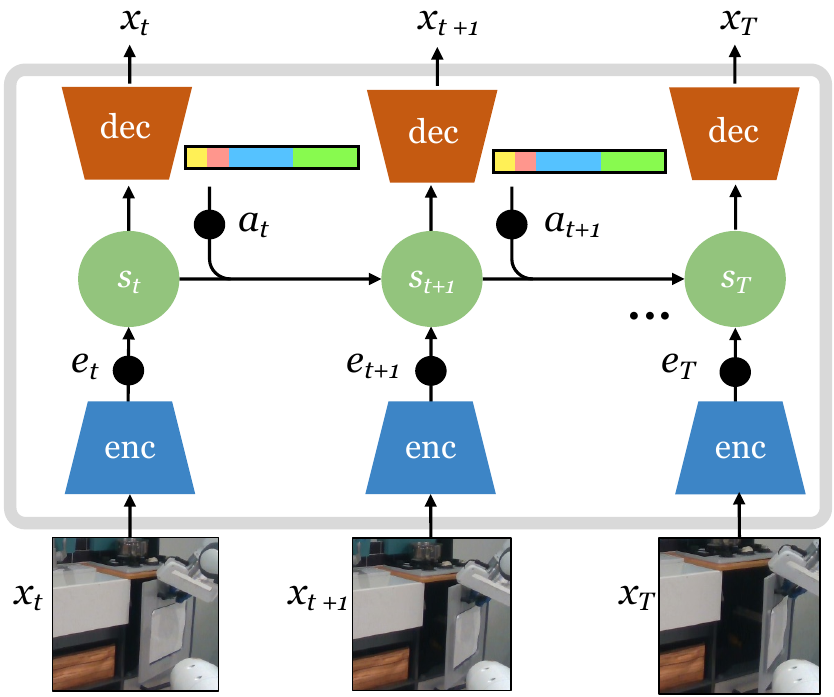}
    \caption{ \small World Model Training: Images and actions are encoded into a learned feature space that has temporal structure, following the approach from~\citet{hafner2020dreamerv2}}
    \label{fig:wm-training}
\end{figure}

\begin{figure*}[t!]
    \centering
    \includegraphics[width=\linewidth]{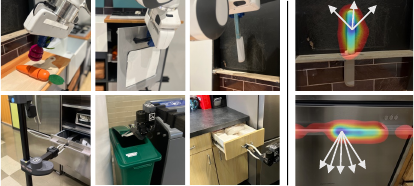}
    \caption{ \small We evaluate \ours on six different real-world manipulation tasks on two different robot systems (shown on the left). On the right, we show a sample of the visual affordances from the visual affordance model $\mathcal{G}_\psi$}
    \label{fig:envs}
\end{figure*}

\section{Background and Preliminaries}

\noindent\textbf{World Models$\quad$}
These are used to learn a compact state space for control given high-dimensional observations like images. The learned states preserve temporal information, which enables effective prediction and planning \cite{ha2018worldmodels, schmidhuber_curiosity, schmidhuber1991curiousmodel}. In this work, we use the model structure and training procedure from Dreamer \cite{hafner2018planet, hafner2019dreamer, hafner2020dreamerv2}, which has the following components: 
\eq{
&\text{encoder:} && e_t=\operatorname{enc_\phi}(x_t) 
&&\text{posterior:} && p(s_t | s_{t-1},a_{t-1},e_t) \\
&\text{dynamics:} && p(s_t | s_{t-1},a_{t-1}) 
&&\text{decoders:} && p(x_t | s_t) , p(r_t| s_t) \\
}
Here $x_t, a_t, r_t$ denote the observation, action, and reward at time $t$, and $s_t$ denotes the learned state space. Note that all these components are parameterized using neural networks. The model is trained by optimizing the ELBO as described in Dreamer, where the learned features are trained to reconstruct images and rewards and are regularized with a dynamics prior. The reward head decoder is not trained if $r_t$ is not provided. For more details, we refer the readers to \citet{hafner2020dreamerv2}. 

\noindent\textbf{Hand-Object Interactions from Human Videos$\quad$}
In this paper, we leverage human videos to learn world models. Throughout the paper, we will refer to a set of visual affordances. These visual affordances comprise of the hand trajectory $h_t$ in image space (normalized to a 0-1 range), and object locations ($o_t)$. We obtain human hand-object information $(h_t, o_t)$ for each frame using the 100 Days of Hands \cite{100doh} detector model, trained on many hours of youtube videos. These can then be used to identify where on the object the hand makes contact $p^g$, and we sample the hand position from a later frame in the video to obtain $p^{pg}$. Here $p^g$ and $p^{pg}$ denote the grasp and post-grasp pixel respectively and specify the visual affordance space.

\section{World Models from Human Videos}

\subsection{Visual Affordances as Actions}

\label{sec:vis_affordance}
One of the key challenges is defining what the actions should be from human videos, most of which just contain image observations. Action information is essential for world models since they are required to learn dynamics and make predictions about the future. Furthermore, we need to define actions in a manner that is \emph{transferable} from the human video domain to robot deployment settings. Following previous work that studies human-to-robot transfer for manipulation \cite{bahl2022human, shao2021concept2robot, chen2021dvd, xiong2021learning, smith2019avid, shaw2022video, sharma2019third, sermanet2016unsupervised, zakka2021xirl}, we use the human hand motion in the videos to inform the action space. This is because we are focused on performing manipulation tasks, and the manner in which humans interact with objects using their hands contains useful information that can be transferred to robot end-effectors. 

\noindent\textbf{Structured Actions from Videos$\quad$} We note that the videos of humans interacting with objects often consist of the hand moving to a point on the object, performing a grasp, and then manipulating the object. 
 After obtaining the grasp pixel $p^g$ and post-grasp pixels $p^{pg}$, using computer vision techniques similar to \cite{hap, hoi, hotspots} from the video clip, we use these to train $\mathcal{G_\phi}$, which distills these labels into a neural network model conditioned on the first frame of the video clip. This model thus learns affordances associated with objects in the scene, by modeling how humans interact with them. This follows the affordances described in \citet{vrb}, but our work can also be combined with other affordance-learning approaches.

\noindent\textbf{Transfer to Robot Scene$\quad$} When dealing with 2D images, there is an inherent ambiguity regarding depth, which is required to map to a 3D point. To overcome this, we utilize depth camera observations to obtain the depth $d^{g}_t$ at the image-space point $p^{g}_t$, and also sample the post-grasp depth $d^{pg}_t$ within some range of the environment surface. This can then be projected into 3D coordinates in the robot frame, using hand-eye calibration, and the robot can attempt to grasp and manipulate objects by moving its gripper to these locations. The affordance action at time $t$ can thus be expressed as $u_t = [p_t, d_t]$, where $d_t$ is the depth corresponding to pixel $p_t$.

\label{sec:hybrid_acts}
\noindent\textbf{Hybrid Action Space$\quad$} While visual affordances help structure the action space to increase likelihood of useful manipulation and allow us to learn from human video, they impose restrictions on the full space of end-effector motion. Hence, we adopt a hybrid action space that has the option to execute both the aforementioned visual affordance, as well as arbitrary end-effector Cartesian actions. We append a mode index to denote which type of action should be executed. This enables the robot to benefit both from the structured pixel-space visual affordance actions and the pre-training data in mode ($m$) 0, and make adjustments using arbitrary end-effector delta actions in mode 1. An action can be described by the following: 

\vspace{-0.2in}

\begin{align}\label{eq:act-def}
    a_t = [m_t, \theta_t, u_t, \Delta y_t] 
\end{align}

Here $m_t$ denotes the mode, $\theta_t$ is the rotation of the gripper, $u_t$ is the image-space action ($u_t$ = $[p_t, d_t]$, where $p_t$ are pixel coordinates in the image and $d_t$ is depth), and $\Delta y_t$ is the Cartesian end-effector action. At a particular timestep, only one out of the image action and Cartesian actions can be executed.  If $m_t = 0$, this corresponds to the affordance mode, and so $p_t$ is executed. If $m_t = 1$, then the robot is operating in the Cartesian control mode, and $\Delta y_t$ is used. Due to our hybrid action space, we can seamlessly switch between training with the visual affordance and Cartesian end-effector action spaces. This allows the robot to leverage the structure from human video and also make adjustments if required using Cartesian actions which are useful for fine-grain control.

\begin{algorithm}[ht!]
\caption{  Human Video Data Training}
\label{alg:passive_process}
\begin{algorithmic}[1]
\REQUIRE Human Video Dataset $\mathcal{D}$
\STATE \textbf{initialize:} World model $\mathcal{W}$, Affordance model $\mathcal{G}$
\STATE Process $\mathcal{D}$ into video clips $C^{0},... C^{T}$
\STATE Obtain grasp $p^g$ and post grasp $p^{pg}$ pixels for each $C^{k}$. 
\STATE Create actions $a_t$ using eq. \ref{eq:act-def}, with mode $m_t = 0$, and randomly sampling depth $d_t$ and rotation $\theta_t$
\STATE Train $\mathcal{G}_\phi(a^{g}, a^{pg} | I^{k}_0)$, where $I^{k}_{0}$ is the first frame of $C^{k}$
\STATE Train $\mathcal{W}$ on trajectory sequences $ \{(I^{k}_0, a^{g}, I^{k}_{t1}, a^{pg}, I^{k}_{t2})\} $
\STATE \textbf{return} $\mathcal{W}, \mathcal{G}$
\\[.7ex]
\end{algorithmic}
\vspace{-0.05in}
\end{algorithm}

\begin{figure*}%
    \centering
    \includegraphics[width=\linewidth]{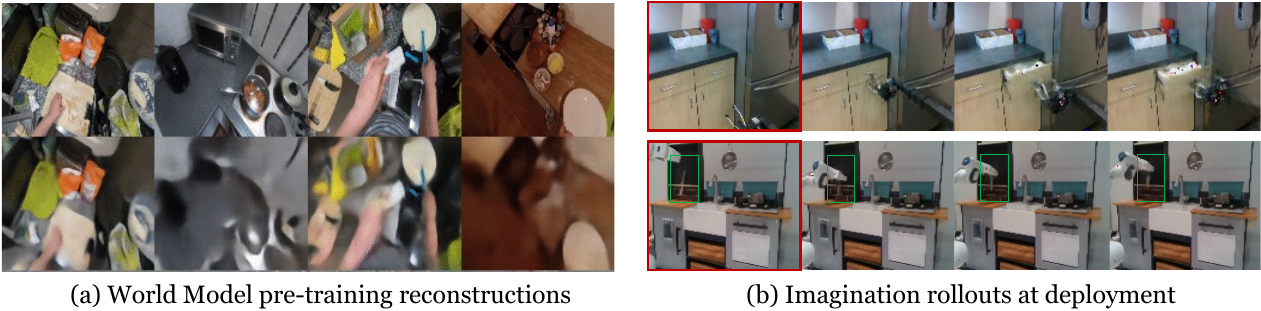}
    \caption{ \small a) World Model pre-training reconstructions on Epic-Kitchens dataset \cite{EPICKITCHENS}. b) Model imagination rollouts for high-reward trajectories. We can see that \ours can imagine plausible and successful trajectories, for both human and robot data. The first image (highlighted in red) is the original observation by the robot.}%
    \label{fig:model-rollouts}%
\end{figure*}

\subsection{Structured Affordance-based World Models for Robotics}

The overall approach is outlined in Alg. \ref{algo:full_method}. We now describe each of the three phases - 1) World model pre-training on human videos, 2) Unsupervised finetuning with robot data, and 3) Robot deployment to perform a task given a goal image.

\noindent\textbf{Training from Passive Human Videos $\quad$} 
 We first use a large set of human videos, obtained from Epic-Kitchens \cite{EPICKITCHENS} to both train the world model $\mathcal{W}$, and obtain the visual affordance model $\mathcal{G_\phi}$. This dataset includes around 50k egocentric videos of people performing various manipulation tasks in kitchens. We first process this dataset into a set of short video clips (around 3 seconds). After obtaining the grasp pixel $p^g$ and post-grasp pixels $p^{pg}$ from the video clip, we convert them to our action space (specified in eq. \ref{eq:act-def}), and train $\mathcal{G_\phi}$, as previously described in section \ref{sec:vis_affordance}. For video clip $k$, let $I^{k}_t$ denote an image frame from the clip at time $t$.  We collect images $I^k_{t_1}$ and $I^k_{t_2}$, where $t_1$ is the time of the grasp, and $t_2$ is when the hand is at $p^{pg}$. $\mathcal{W}$ is then trained on the trajectory sequences: 

 \begin{align}
     \{I^{k}_0, a^g I^{k}_{t_1}, a^{pg} I^{k}_{t_2}\}
 \end{align}
 
 This procedure is outlined in Alg. \ref{alg:passive_process}. As described in section \ref{sec:hybrid_acts}, there are two modes for the actions - either in pixel space or end-effector space. In order to train on human videos, we consistently set $m_t = 0$ and thus use the image space actions. Since image depth and robot rotation information are not present in the video, we randomly sample values for these components. We include visualizations of the world model predictions on the passive data in Figure \ref{fig:model-rollouts}, and see that the model is able to capture the structure of the data. 

\begin{table*}[t!]
\centering
\resizebox{0.9\linewidth}{!}{%
\centering
\begin{tabular}{l@{\hskip 0.2in} c@{\hskip 0.2in} c@{\hskip 0.2in} c@{\hskip 0.2in} c@{\hskip 0.2in} c@{\hskip 0.2in} c@{\hskip 0.2in} c}
\toprule
 & \textbf{Cabinet} & \textbf{Veg} & \textbf{Knife} & \textbf{Drawer}  & \textbf{Dishwasher}  & \textbf{Can} & \textbf{Average} \\
\midrule
\multicolumn{7}{l}{\textit{No world model}:}\vspace{0.4em}\\
\texttt{BC-Affordance}  & 0.32 & 0.48 & 0.16 & 0.56 & 0.20 & 0.44  & 0.36 \\
\texttt{BC-Pix} & 0.16 & 0.40 & 0.00 & 0.24 & 0.08 & 0.12 & 0.17  \\
\midrule
\multicolumn{7}{l}{\textit{No human-centric affordance-based actions}:}\vspace{0.4em}\\
\texttt{MBRL-single \cite{hafner2020dreamerv2}}  & 0.00 & 0.28 & 0.20 & 0.00 & 0.04 & 0.00 & 0.09 \\
\texttt{MBRL-Pix-single}  & 0.52 & 0.36 & 0.16 & 0.00 & 0.04 & 0.04 & 0.19\\
\midrule
\multicolumn{7}{l}{\textit{No pre-training from human videos}:}\vspace{0.4em}\\
\texttt{MBRL-Affordance-single} & 0.68 & 0.16 & 0.40 & 0.84 & 0.20 & 0.36 & 0.44 \\
\texttt{MBRL-Affordance-joint} & 0.12 & 0.36 & 0.36 & 0.08 & 0.20 & 0.04 & 0.19\\
\midrule
\texttt{\textbf{\ours}}  & 0.84 & 0.76 & \textbf{0.72} & 0.92 & \textbf{0.84} & \textbf{0.68} & \textbf{0.79} \\ 
\texttt{\textbf{\ours-single}}  & \textbf{0.88} & \textbf{0.80} & 0.60 & \textbf{0.96} & 0.68 & 0.56 & 0.75 \\
\midrule
\bottomrule
\end{tabular}}
\caption{ \small Success rates of \ours and baselines on six different manipulation tasks, over 25 trials. }
\label{tab:main}
\end{table*}

\begin{algorithm}[t!]
\caption{ Overview of \ours}
\label{alg:method}
\begin{algorithmic}[1]
\STATE Get $\mathcal{W}, \mathcal{G}$ = Human Video Data \textbf{Pre-Training} (Alg. \ref{alg:passive_process})
\STATE \textbf{Finetuning}: Query $\mathcal{G}$ for $N_0$ iterations to collect robot dataset $\mathcal{R_D}$ to train $\mathcal{W}$.
\STATE \textbf{Task Deployment: } (Given goal $I_g$)
\STATE Rank trajs in $\mathcal{R_D}$ using $I_g$. Fit GMM $g$. 
\FOR{traj 1:K}
\STATE Query $N$ proposals from $\mathcal{G}$, $\{{a^{g}, a^{pg}} \}_{1..N}$
\STATE Query $M$ proposals from $g$
\STATE Select best proposal using CEM through $\mathcal{W}$ 
\STATE Execute on the robot to reach $I_g$
\ENDFOR
\\[.7ex]
\end{algorithmic}
\vspace{-0.05in}
\label{algo:full_method}
\end{algorithm}

\noindent\textbf{Finetuning with Robot Data$\quad$}  To use the world model $\mathcal{W}$ for control, we need to collect some in-domain robot data for finetuning. We do so by running the visual affordance model $\mathcal{G}$ to collect a robot dataset $R_D$, which is then used to train $\mathcal{W}$. We emphasize that this step does not require \emph{any} supervision in the form of task rewards or goals. Hence, we can collect data from diverse tasks in the finetuning step. We see in Fig. \ref{fig:rollout-pt} that \ours enables the world model to pick up on the salient features of the robot environment very quickly as compared to models that do not use pre-training on human videos.

\noindent\textbf{Task Deployment $\quad$}
After the world model has been fine-tuned on robot domain data, it can be used to perform tasks specified through goal images. The procedure for doing so is outlined in the Task Deployment section in Alg. \ref{algo:full_method}. We collect two sets of action proposals. The first set is obtained by querying the visual affordance model $\mathcal{G}$ on the scene. We also want to leverage our knowledge of trajectories in $\mathcal{R_D}$ that reach states close to the goal. For this, we create a second set of proposals by fitting a Gaussian Mixture Model to the top trajectories in $\mathcal{R_D}$ and sampling from it. 
We then use the world model to optimize for an action sequence using the standard CEM approach \cite{cem}, where the initial set of plans is set to be the combined set of action proposals. Ranking the trajectories in $\mathcal{R_D}$ and running CEM requires rewards, and we can obtain this by measuring the distance to the goal in the world model feature space:

$$ r_t = \text{cosine}(f_{\mathcal{W}}(I_{g}) , f_t)  $$

where $f_t$ is the world model feature, and $f_{\mathcal{W}}$ is the learned feature space of the model. For ranking trajectories in $\mathcal{R_D}, f_t = f_{\mathcal{W}}(I_{k})$ for image $k$ in the dataset. For planning, $f_t$ corresponds to the predicted feature state. In our experiments, we use cosine distance to goal in the feature space from \citet{r3m} to provide reward for model-free baselines, since they do not have a model, and so we also add this term to our reward by training a reward prediction head to get feature space \cite{r3m} distance to goal from $f_t$. In our experiments we run an ablation where we use only the world model feature space, and find that performance for our approach is about the same. 

\section{Experimental Setup}

\subsection{Environments}

Our real-world system consists of two different robots, evaluated over six tasks. Firstly, we use the Franka Emika arm, with end-effector control. This robot acts in a play kitchen environment with multiple tasks that mimic a real kitchen. Specifically, the robot needs to open a cabinet, pick up one of two toy vegetables from the counter and lift a knife from a holder. Note that the knife task is very challenging as it requires fine-grained control from the robot. In order to test \ours in the wild we also deploy it on a mobile manipulator, the Stretch RE-1 from Hello-Robot. This is a collaborative robot designed with an axis-aligned set of joints and has suction cups as fingertips. We run this robot in real-world kitchens to perform different tasks, including opening a dishwasher, pulling out a drawer and opening a garbage can. The garbage can task is challenging as the area for the robot to grasp onto is quite small. We show images of the environments in Figure~\ref{fig:envs}.

\subsection{Baselines and Ablations}

In order to compare different aspects of \ours, we run an extensive of baselines and ablations. All world-model-based approaches directly use code from Dreamer \cite{hafner2020dreamerv2}.

\begin{itemize}
    \item \texttt{MBRL-Affordance: } An important contribution of \ours is pre-training on human videos. This baseline is similar to \ours but does not use any human video pre-training, allowing us to test our hypothesis that using human video is important for learning a generalizable world model. 
    \item \texttt{MBRL-Pix: } Secondly, we would like to test how much the affordance action space helps the robot. This approach uses the same world model control procedure as \ours, but does not sample actions using the visual affordance, $\mathcal{G}_\psi$. Instead grasp and post grasp locations are randomly sampled from an image crop around the object.
    \item \texttt{MBRL: } This baseline further removes structure from the action space, and only uses cartesian end-effector actions, without any pixel-space structure, thus $m_t = 1$ (described in section \ref{sec:hybrid_acts}) for every timestep $t$. In order to help with sample efficiency, we use a simple heuristic to bootstrap this baseline: we initialize the robot each episode near the center of the detected object, using Detic \cite{detic}, a state-of-the-art object detector. 
    \item  \texttt{BC-Affordance: } We would like to test if using world models is critical to performance, or if a simple behavior cloning approach can be effective. This baseline employs a filtered-behavior cloning \cite{peng2019advantage, rwr} strategy, in which the top trajectories based on reward (in our case distance to goal) are selected. Since there is no learned world model, we use distance in the feature space from the R3M model \cite{r3m}. After selecting the top trajectories, we fit a gaussian mixture model and sample from it to obtain actions. These are in the same visual affordance action space used by our approach.
    \item \texttt{BC-Pix: } Uses behavior cloning in the same way as \texttt{BC-Affordance}. The only difference is the action space - this approach randomly samples locations and does not use $\mathcal{G}_\psi$ to obtain actions.
\end{itemize}

\begin{figure*}[t!]
    \centering
    \includegraphics[width=\linewidth]{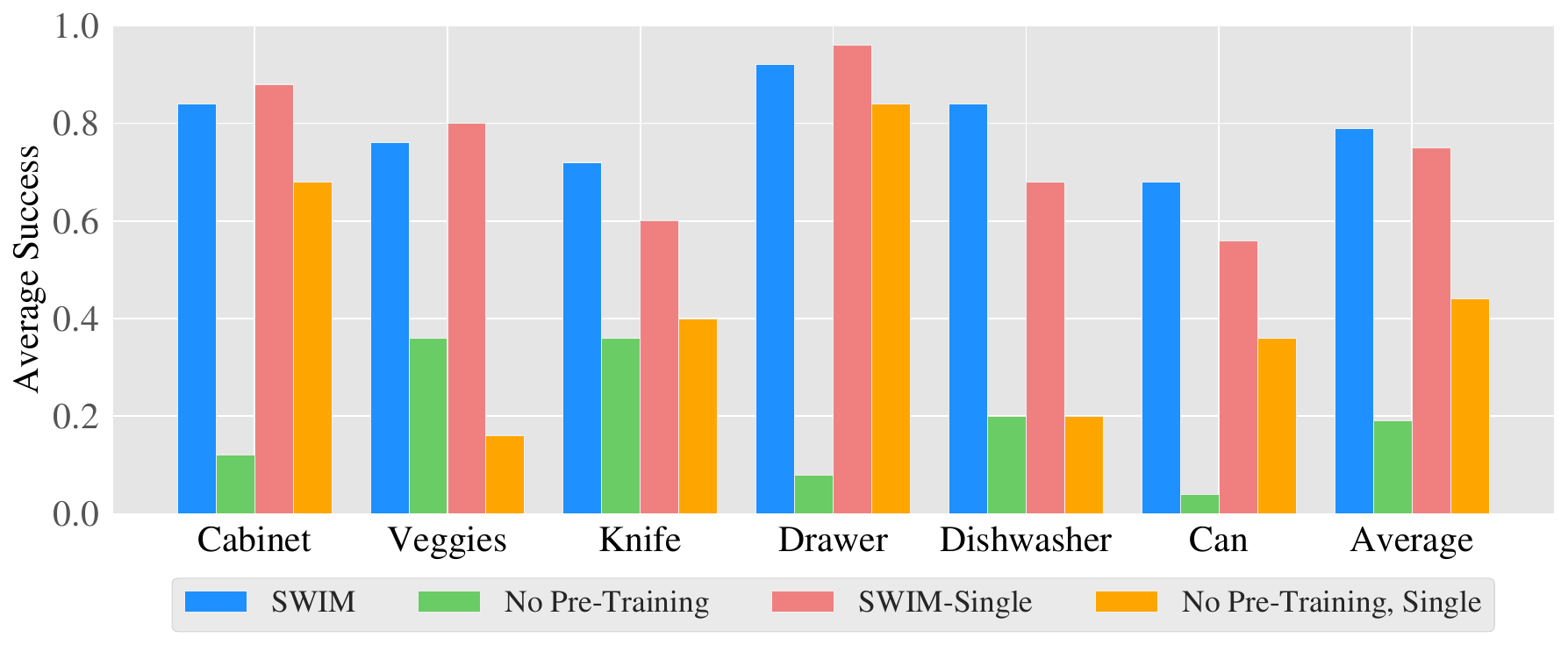}
    \caption{ \small Comparison of \ours and \texttt{MBRL-Affordance} for both the single task and jointly trained model. We see a large drop in success when removing pre-training on human videos, especially when dealing with diverse robot tasks.}
    \label{fig:pt-bar}
\end{figure*}

\subsection{Implementation details}

\noindent\textbf{Human Video Data Pre-trainig $\quad$} In order to pre-train the world model on human videos, we use the Epic-Kitchens \cite{EPICKITCHENS} dataset. The dataset is divided into many small clips of humans performing semantic actions. We use the 100 Days of Hands \cite{100doh} detector to find when an object has been grasped and find post grasp waypoints. Around 55K such clips are used to train the world model. Since we do not have depth or 3D information available, we randomly sample $\theta_t$ and the depth component of the image space action $p_t$. 

\noindent\textbf{Affordance Model $\quad$} We show some qualitative examples of the affordances of the human-affordance model ($\mathcal{G}_\psi$) we use in Figure~\ref{fig:envs}. This model has a UNet style encoder-decoder architecture, with a ResNet18 \cite{resnet} encoder. The final output of the model is $h_t$ and $g_t$, where $g_t$ is a set of keypoints obtained from a spatial softmax over the network's heatmap outputs, representing the grasp point, and $h_t$ is the post-grasp trajectory of the detected hand.

\noindent\textbf{World Model $\quad$} We use the world model from \citet{hafner2020dreamerv2}. However, in order to handle high-dimensional image inputs we employ NVAE \cite{vahdat2020nvae} as a stronger visual encoder. While not necessary to train the reward model $q_r$ when finetuning, we empirically found that it added stability to the filtering setup a test time. We leave distilling the latent features into a neural distance function as future work. 

\noindent\textbf{Robot Deployment Setup$\quad$} To capture videos and images we use an Intel Realsense D415, to get RGBD images. For each task we collect either 25 or 50 iterations of randomly sampled actions (in human-affordance, random image or Cartesian space), which takes about 30 minutes, finetuning the model on collected data. We obtain feature distance w.r.t. to image goals using the ResNet18 encoder from \citet{nair2022r3m}. We sample around 2K action proposals and use the output of $\mathcal{W}_\phi$ to prune these. The model outputs are then evaluated (25 times). A human measures success based on a pre-defined metric (i.e. the cabinet should be fully open, etc).

\section{Results}
In our experiments we ask the following questions (i) Can we train a \emph{single world model jointly} with data coming from diverse tasks? (ii) Does training the world model on \emph{human video} data help performance? (iii) How important is our structured action space, based on human visual \emph{affordances}? (iv) Are \emph{world models} beneficial for learning manipulation with a handful of samples? (v) Can our approach \emph{continually improve} performance with iterative finetuning?

\begin{figure*}[t!]
\centering
\begin{subfigure}[b]{0.329\linewidth}
    \includegraphics[width=\linewidth]{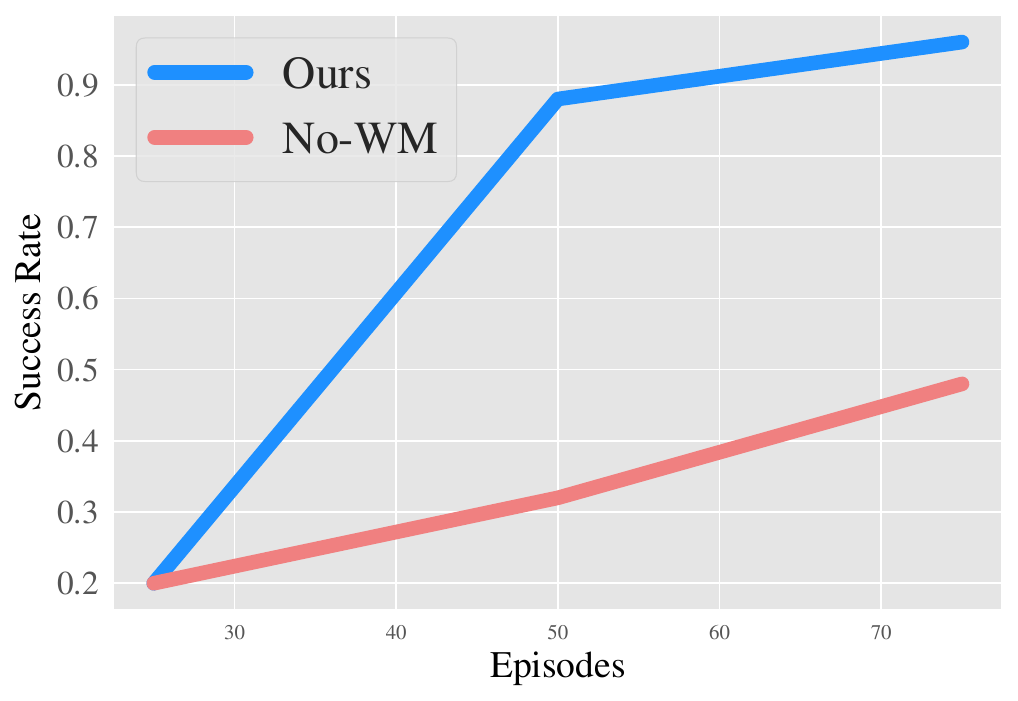}
    \caption{\small Cabinet}
\end{subfigure}
\begin{subfigure}[b]{0.329\linewidth}
    \includegraphics[width=\linewidth]{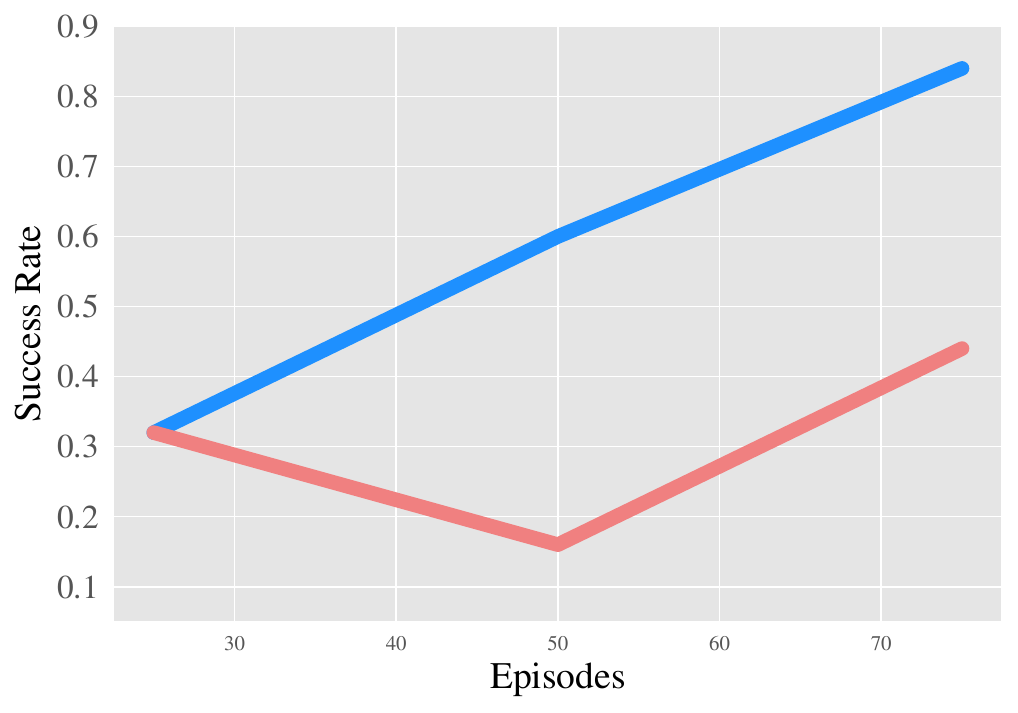}
    \caption{\small Knife}
\end{subfigure}
\begin{subfigure}[b]{0.329\linewidth}
    \includegraphics[width=\linewidth]{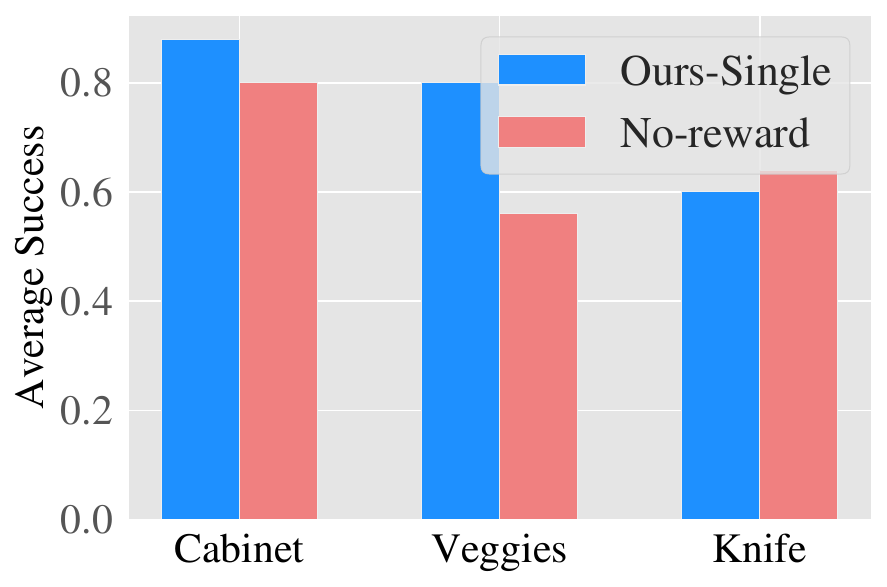}
    \caption{\small No-Reward}
\end{subfigure}
\caption{\small Continuous improvement (a-b): We see that \ours continues to improve, achieving high success. (c) Ablating the need for external feature space goal distance at test time.}
\vspace{-0.1in}
\label{fig:improvement}
\end{figure*}

We present a detailed quantitative analysis of various approaches in Table \ref{tab:main}. We see that across environment settings and robots \ours achieves an average success rate of about 80\% when using joint models (trained separately for Franka and Hello robot tasks). We also observe strong performance when \ours is trained on individual tasks, getting an average success of 75\%, compared to the next best approaches which only get around 40\% success.

\begin{figure}[ht]
    \centering
    \includegraphics[width=\linewidth]{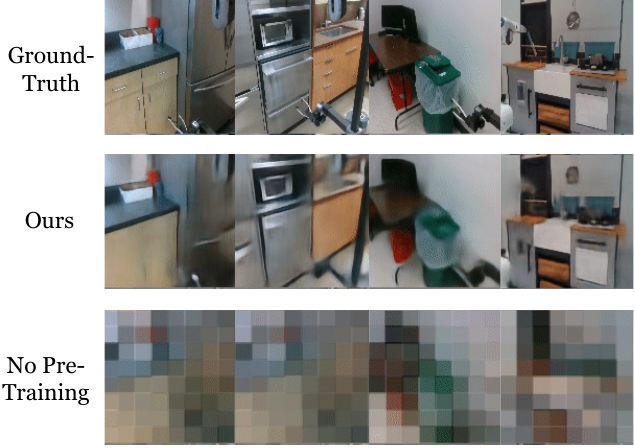}
    \caption{\small Image reconstruction using world model features in early training stages for \ours and \texttt{MBRL-Affordance} (which has no pre-training), showing that \ours can effectively transfer representations from human videos. Note that for our experiments we use models trained to convergence.  }
    \label{fig:rollout-pt}
    \vspace{-0.15in}
\end{figure}

\noindent\textbf{Joint World Model$\quad$} A big benefit of \ours is that it can deal with different sources of data. \texttt{\ours-single} employs a model trained individually for each task. We see that overall the performance improves when sharing data, from the last two rows of Table \ref{tab:main}. This is likely because there are some similarities across tasks that the model is able to capture. We find this encouraging and hope to scale to more tasks in the future. Further, we see that for the best baseline, \texttt{MBRL-Affordance-single}, using all the data jointly to train the world model leads to a major drop in performance (from 44\% to 19\%). We show this visually in the bar chart in Figure~\ref{fig:pt-bar}, where the effect of pre-training a model on human videos is amplified when dealing with all of the data from all the tasks. This shows that when dealing with a large set of tasks and diverse data, it is crucial to incorporate human-video pre-training for better performance and generalization. Hence we do not run joint world model experiments for \texttt{MBRL} and \texttt{MBRL-Pix} since the performance is already quite low (around 10 - 20 \%) when the model is trained just on single task data.

\noindent\textbf{Human Video Pre-Training$\quad$} As noted in the previous section, pre-training on human videos is \emph{critical} to being able to effectively train joint world models on multi-task data, as seen in Figure ~\ref{fig:pt-bar}. For \texttt{MBRL-Affordance} we saw that in many cases the model collapses quickly to a sub-optimal control solution when trained on multiple tasks jointly. To investigate this further, we visualize the image reconstructions from $\mathcal{W}$ within the first minute of training and find that the outputs of \ours were already very realistic, as compared to those of \texttt{MBRL-Affordance}. This can be seen in Figure~\ref{fig:rollout-pt}, where the outputs of \texttt{MBRL-Affordance} are very pixelated while those of \ours already capture important aspects of the ground truth, indicating the usefulness of pre-training on human videos.

\noindent\textbf{Human-Affordance Action Space$\quad$} How does the choice of action space affect performance? For this we compare the (single task) model based and BC approaches separately. Comparing \texttt{MBRL-single} and \texttt{MBRL-Affordance-single} in Table \ref{tab:main},  we can see that there is a clear benefit in using structured action spaces, with over 5X the success compared to cartesian end effector actions. This fits our hypothesis, as it is very difficult for methods that use low-level actions to find successes in a relatively small number of interaction trajectories. The few successes that \texttt{MBRL-single} does see are due to the initialization of the robot close to the object using Detic. Furthermore, we note that this benefit is not simply because the affordance actions are in image space. In both the filtered-BC and world model case, the success rate with the affordance action space is roughly \emph{double} than that of acting in pixel space, where target locations are sampled from a random crop around the object.  This shows that picking the right action space~\textit{and} acting in a meaningful way to collect data can bootstrap learning and lead to efficient control.

\noindent\textbf{Role of World Model$\quad$} How important is using a world model, and can we achieve good performance by just using the affordance action space? From Table~\ref{tab:main}, we see that the average performance of \texttt{BC-Affordance} is not too far behind that of \texttt{MBRL-Affordance-single}. However, without a world model, the controller cannot leverage \emph{multi-task} data, both for the pre-training stage to use human videos and for learning the shared structure across multiple robot domains by training a joint world model. Due to these critical reasons discussed previously, \ours outperforms the best filtered-BC approach by more than a factor of 2.

\noindent\textbf{Continual Improvement$\quad$} Next, we investigate if \ours can keep improving using the data that it collects when planning for the task. Since the world model can learn from all the data, we want to test if it can improve its proficiency on the task. Thus, after evaluating $\mathcal{W}$ once, we retrain on the newly collected data (as well as the old data), and re-evaluate the model. We present the learning curves in Figure \ref{fig:improvement}. We see that \ours is able to effectively improve performance, and achieves success of over 90\%, which is far better than the performance of \texttt{BC-Affordance} even after continual training. This is an encouraging sign that \ours can scale well since it can keep improving its performance with more data to continually learn. In the future, we hope to not only continually finetune, but also add new tasks and settings. 

\noindent\textbf{Reward Model$\quad$} In Figure~\ref{fig:improvement} c) we examine the effect of removing the reward prediction module on planning and find that only using distance in world model feature space is fairly competitive with using both the feature distance and predicted reward. We hypothesize that for the veggies task, it was harder to estimate reward accurately because the free objects tend to move around a lot during training, thus it might take more samples to learn consistent features.

\section{Discussion and Limitations}
In this paper, we present \ours, a simple and efficient way to perform many different complex robotics tasks with just a handful of trials. We aim to build a single model that can learn many tasks, as it holds the promise of being able to continuously learn and improve. We turn to a scalable source of useful data: human videos, from which we can model useful interactions. In order to overcome the morphology gap between robot and human videos, we create a structured action space based on human-centric affordances. This allows \ours to pre-train a world model on human videos, after which it is fine-tuned using robot data collected in an unsupervised manner. The world model can then be deployed to solve manipulation tasks in the real world. The total robot interaction samples for the system can be collected in just 30 minutes.  Videos of \ours can be found at \texttt{\url{https://human-world-model.github.io}}. While \ours provides a scalable solution and shows encouraging results, some limitations are in the types of actions and tasks that can be performed, as they currently only include quasi-static setups. In future work, we hope to explore different action parameterizations and other types of manipulation tasks. We also hope to scale to many more tasks in order to build a truly generalist agent that can keep on getting better by learning from both passive and active data. 

\vspace{0.1in}
\noindent\textbf{Acknowledgement}$\quad$We thank Shagun Uppal and Murtaza Dalal for feedback on early drafts of this manuscript. This work is supported by the Sony Faculty Research Award and ONR N00014-22-1-2096.

\bibliographystyle{plainnat}
\bibliography{references}

\newpage
\section*{Appendix}
\subsection{Videos}

Videos of our results can be found at: \texttt{\url{https://human-world-model.github.io}}, or in the zip folder. 

\subsection{Implementation Details}

\subsubsection{Robot Setup}

We use two robots: Franka Emika and Stretch RE1 from Hello-Robot. Both robots are controlled in end-effector space as well as a rotation (roll for the Stretch, and roll + pitch for the Franka). The Franka roll and pitch, as well as the Stretch roll are sampled from $[0, -\frac{\pi}{4}, -\frac{\pi}{2}]$ (randomly at first). The robots run open loop trajectories. The camera observations are coming from D415 Intel Realsense RGBD cameras. We use a low-level impedance controller for the Franka to reach the desired high level actions.

\subsubsection{Tasks and Environments} Our setup consists of six tasks, three (veggies, knife and cabinet) of which are in a Play Kitchen from \citet{ebert2021bridge}, and we have three in the wild tasks that involve opening the dishwasher, lifting the garbage can handle or pulling out a drawer. These are everyday tasks that we found. Videos of each task can be seen at \texttt{\url{https://human-world-model.github.io}}.

\subsubsection{Data Collection} We perform data collection by executing the affordance model $\mathcal{G}_\psi$ at first, and then set mode $m_t = 1$, and use $\Delta y_t$ as Cartesian end-effector deltas. These are sampled from $\mathcal{N}(0, 0.05)$. We sample from $\mathcal{G}_\psi$ in the following manner: we obtain the 2D pixel from the model, and find the depth at that point. For the grasp part of the affordance, we simply pass this depth to our controller (which has hand-eye calibration). For post grasp trajectory, we sample a depth with $d$ as the center, with a bias towards moving away from the surface (as we usually have a wall right behind the object, or the depth camera). For each of the baselines, we use the underlying action space to sample actions, and append $\Delta y_t$ to the end. Our trajectory, during the robot sampling stage, consists of 3-4 actions with $m_t = 0$, and 6-10 actions with $m_t = 1$. The overall data collection process takes about 25 to 45 minutes depending on how long resets takes. 

\subsubsection{Human Videos}

Our human video dataset is obtained from Epic-Kitchens \cite{EPICKITCHENS}. We take semantically pre-anotated action clips, and apply the 100 Days of Hands (100 DoH) \cite{100doh} hand-object model to get annotations for when and where the contact happened, and how the hand moved post contact, all in normalized (0, 1) pixel space. To obtain the contact points, we use a similar pipeline to \citet{hoi}, where we find the intersection of the hand bounding box and the interacted object's bounding box, and look for skin outline in that region. We use a skin segmentation (similarly to \citet{hoi}) to get the external grasp points. We obtain about 55K such clips to train on. Each sequence is of length $4$, with $m_t = 0$ for all $t$. For the rotation and the depth values, we randomly sample these values during training, from one of the feasible rotations, or within 50cm of the environment surface respectively. We train a ResNet18 based encoder-decoder architecture for our grasp point prediction. We perform a spatial softmax on the decoder deconvolutional output to obtain the grasp keypoints. The post-grasp trajectory head is a Transformer \cite{vaswani2017attention} with 6 self-attention layers that have 8 heads, inspired from \citet{hoi}.

\subsubsection{World Model}

Our world model architecture is the same as that of \citet{hafner2020dreamerv2}, excluding the visual encoders or decoders. We do not tune any of the world model hyper-parameters, and use the default Dreamer\cite{hafner2020dreamerv2} settings. We use the NVAE \cite{vahdat2020nvae, babaeizadeh2021fitvid} encoder and decoder used in FitVid \cite{babaeizadeh2021fitvid} to better handle high dimensional image prediction. We use only one cell per block instead of two, due to GPU memory restrictions and to train with larger batch sizes. We do not have any residual connections between the encoder and the decoders, to force the latent of the world model, $z$, to be an information bottleneck. The dimension of $z$ is 650 (the determisitic component of the RSSM\cite{hafner2020dreamerv2} is size 600, and the stochastic component is size 50). The model is trained in Tensorflow, and each image is of the size 128x128x3. In the experiments that use reward prediction, we reqgress $q_r$ (the reward decoder) to the distance to goal in the space of R3M \cite{r3m} features (the ResNet18 \cite{resnet} version) of the weights. The reward predictor network consists of a 2 layer MLP with 400 hidden units which takes the world model feature $z$ as input.

\subsubsection{Baselines}

Every baseline that uses a world model uses the same code as \ours, with either a different pre-training setup or different action space. 

\begin{itemize}
    \item \texttt{MBRL-Affordance}:  This is the same exact setup as \ours in terms of the world model and the execution of the affordance model, but we do not use any pre-trained weights when training on robot data. 
    \item \texttt{MBRL-Pix}: The action type is the same as 
    \texttt{MBRL-Affordance}, but the pixel locations are chosen at random, and not from the human-centric affordance model. The actions are sampled uniformly in the 2D crop around the object. 
    \item \texttt{MBRL: } Here all of the actions are with $m_t = 1$. 
    \item  \texttt{BC-Affordance: }   This is a filtered-behavior cloning \cite{peng2019advantage, rwr} strategy. We rank trajectories based on distance in R3M \cite{r3m} space to goal. We fit a Gaussian Mixture Model with 2 centers to the top actions, and sample from those, at execution time. 
    \item \texttt{BC-Pix: } We fit a GMM top trajectories just like \texttt{BC-Affordance}. The sampling space is uniform in the crop around the object. 
\end{itemize}

\subsubsection{Training, Finetuning and Deployment}
For training the world model, $\mathcal{W}_\phi$, in each iteration we train on 100 batches of data, where each batch consists of an entire trajectory sequence. These sequences are of length 2, 3 and 10 for the human video, hello robot and franka robot settings respectively. We first train a model on the human data for about 6000 such iterations with a batch size of 80, which takes about 96 hours on a single RTX 3090 GPU (using 24GB of VRAM). We then finetune this model for 300 epochs on robot data for the joint model, and 200 iteration for the single-task models using a batch size of 24, on a RTX 3090, which takes about 3-4 hours of training. The batch size for robot data is smaller because the model needs to deal with longer sequences consisting of hybrid actions (both the affordance actions and cartesian end effector actions). For the continual learning  experiments we subsequently train on the aggregated datasets for an additional 50 iterations. 
When deploying the model to perform a task, we use CEM for planning at the beginning of the trajectory, and then execute the optimized action sequence in an open-loop manner. We use 3 iterations of CEM, amd 2000 action proposals. Further, in all our experiments, we fix $\mathcal{M}$ = 1400 and $\mathcal{N}$ = 600 ($\mathcal{M} \text{ and } \mathcal{N}$ are defined in Alg. 2), for fixing the ratio of biasing the proposals sent to the model for planning.

\subsubsection{Evaluation}

We evaluate our world model's outputs by executing the trajectory it outputs in the real world using open-loop control. We use goal images that indicate objects are manipulated in specific ways, for example an open cabinet, vegetables picked up and in the air, the knife should be lifted up, the drawer pulled out, the garbage can and dishwasher opened. We evaluate for each method/ablation 25 times, presenting the average.

\subsubsection{Codebases}

We use the following codebases: 

\begin{itemize}
    \item \url{https://github.com/danijar/dreamerv2} \cite{hafner2020dreamerv2} for the world model code 
    \item \url{https://github.com/ddshan/hand_object_detector} for the  100DoH model \cite{100doh}
    \item \url{https://github.com/epic-kitchens} for Epic-Kitchens \cite{EPICKITCHENS} processing
    \item \url{https://github.com/facebookresearch/r3m} for R3M \cite{r3m} model
    \item \url{https://github.com/facebookresearch/fairo/tree/main/polymetis} \cite{Polymetis2021} for the end-effector control code for the Franka 
    \item \url{https://github.com/orgs/hello-robot/repositories} for Stretch RE-1 
    
\end{itemize}

\end{document}